\documentclass[runningheads]{llncs}
\usepackage[T1]{fontenc}
\usepackage{graphicx}
\usepackage{multirow}
\usepackage{booktabs}
\usepackage{float}

\usepackage{caption}
\usepackage{subcaption}

\usepackage{amsmath}
 \usepackage{amssymb}

\begin{document}
\title{Improving Disease Detection from Social Media Text via Self-Augmentation and Contrastive Learning}
\titlerunning{Improving Disease Detection from Social Media Text}
%
\author{Pervaiz Iqbal Khan\inst{1,2}\orcidID{0000-0002-1805-335X} \and
Andreas Dengel\inst{1,2}\orcidID{0000-0002-6100-8255} \and
Sheraz Ahmed\inst{1}\orcidID{0000-0002-4239-6520}}

\authorrunning{Pervaiz et al.}
%
\institute{German Research Center for Artificial Intelligence (DFKI), Kaiserslautern, Germany \and
RPTU Kaiserslautern-Landau, Germany
\email{\{pervaiz.khan,andreas.dengel,sheraz.ahmed\}@dfki.de}\\
}
\maketitle              
\begin{abstract}

Detecting diseases from social media has diverse applications, such as public health monitoring and disease spread detection. While language models (LMs) have shown promising performance in this domain, there remains ongoing research aimed at refining their discriminating representations. In this paper, we propose a novel method that integrates Contrastive Learning (CL) with language modeling to address this challenge.
Our approach introduces a self-augmentation method, wherein hidden representations of the model are augmented with their own representations. This method comprises two branches: the first branch, a traditional LM, learns features specific to the given data, while the second branch incorporates augmented representations from the first branch to encourage generalization. CL further refines these representations by pulling pairs of original and augmented versions closer while pushing other samples away.
We evaluate our method on three NLP datasets encompassing binary, multi-label, and multi-class classification tasks involving social media posts related to various diseases. Our approach demonstrates notable improvements over traditional fine-tuning methods, achieving up to a $2.48\%$ increase in F1-score compared to baseline approaches and a $2.1\%$ enhancement over state-of-the-art methods.

\end{abstract}
\section{Introduction}
Depression is a widespread mental disorder affecting approximately 280 million individuals globally. Tragically, over 700,000 people per year commit suicide due to depression, making it the fourth leading cause of death among individuals aged 15 to 29\footnote{https://www.who.int/news-room/fact-sheets/detail/depression}. The condition significantly impacts both personal and professional lives. Alarmingly, reports published in August 2021 revealed that 1.6 million individuals in England were on waiting lists for mental health support, with approximately 8 million unable to access treatment from specialists due to not meeting the criteria for treatment \cite{garg2023annotated}.

The utilization of the internet and social media platforms such as Reddit\footnote{https://www.reddit.com/} and X\footnote{https://twitter.com} has surged in recent years. Individuals use these platforms not only to provide feedback on services and products but also to share emotions and health-related information. Analyzing these emotions is crucial for the early detection of depression and other diseases. Early detection can mitigate losses by facilitating better resource planning and the establishment of health-related policies. This prompts the question: can social media content aid in disease detection? Natural language processing (NLP) offers an effective means to analyze emotions from text. However, the task of disease detection poses challenges, as disease words and emotions are sometimes used figuratively rather than literally. Overcoming this challenge necessitates learning appropriate word representations based on their context.

Transformer-based models such as BERT \cite{devlin2018bert}, RoBERTa \cite{liu2019roberta}, and XLNet \cite{yang2019xlnet} excel at learning suitable representations and have achieved state-of-the-art (SotA) performance in various NLP tasks. These models are pretrained on large amounts of unlabeled text data using auxiliary tasks such as Masked Language Modeling and Next Sentence Prediction. They can then be fine-tuned on labeled datasets to further enhance representations specific to the downstream task. Various approaches \cite{khan2022improving,khan2022novel,khan2023unique}. These methods inject perturbations into the embedding matrix, hidden representations, and parameter-space of the models using techniques such as Gaussian noise and the Fast Gradient Sign Method \cite{goodfellow2014explaining}.

In this paper, we propose a novel and effective approach called self-augmentation to generate adversarial examples during the fine-tuning process of these models, thereby enhancing their text representation capabilities. The model is fine-tuned on two streams: the first stream is fine-tuned conventionally by minimizing the classification loss, while the second stream augments its intermediate states with extracted representations from the first stream and then completes its fine-tuning pass. Contrastive loss is then employed to further align the intermediate representations of both streams to improve representations. The overall fine-tuning objective is to minimize two classification losses and one contrastive loss.

The main contributions of this paper are summarized as follows:
\begin{itemize}
    \item We conducted experimentation on three public social media text datasets, spanning binary-class, multi-class, and multi-label classification tasks. Our approach achieved SotA results on these datasets, demonstrating the effectiveness of our proposed method in real-world scenarios. Additionally, we presented a thorough analysis of the experimental results, including visualizations of representations and ablation studies, to provide insights into the performance and behavior of our approach.
    \item We proposed a novel method called self-augmentation, which generates adversarial examples to improve text representation learning of LMs. By jointly training clean and adversarial examples, our approach enhances the discriminative features learned by LMs, leading to improved performance on downstream tasks. This contribution addresses the need for robust and generalizable text representations in NLP applications, particularly in the context of social media data.
    \item We incorporated contrastive learning (CL) method to further enhance the representation of text. By leveraging CL, our approach encourages the model to learn semantically meaningful representations by maximizing the agreement between similar instances and minimizing the agreement between dissimilar instances. This contribution enhances the generalization capabilities of our proposed method, leading to more accurate predictions.
\end{itemize}
\section{Related Work}
This section provides an overview of existing approaches in the literature for disease detection from social media platforms such as Reddit, X, etc. We present various methodologies ranging from traditional machine learning and deep learning techniques to more recent advancements such as adversarial training and contrastive learning with an aim to provide a comprehensive understanding of the domain.
\subsection{Disease Detection from Social Media}
Karisani et al. \cite{karisani2018did} curated a dataset comprising 7,192 English Tweets focusing on health-related discussions and disease-related terms. They proposed the WESPAD approach, which involved partitioning and distorting embeddings to enhance generalization on unseen data and address data scarcity issues.

Jiang et al. \cite{jiang2018identifying} introduced another English Tweets dataset with 12,331 labeled examples, utilizing non-contextual embeddings and Long Short-Term Memory Networks (LSTMs) \cite{hochreiter1997long} to classify tweets into personal experience tweets (PET) or non-PET. They also incorporated feature engineering techniques, improving classification results compared to traditional models like support vector machines, decision trees, and k-nearest neighbors.

Iyer et al. \cite{iyer2019figurative} presented FeatAug+, a method that calculated word similarities using an external data source, Sentiment140. These similarity scores were concatenated with language-based features and fed into a convolutional neural network (CNN)-based classifier.

Biddle et al. \cite{biddle2020leveraging} expanded the PHM-2017 dataset \cite{karisani2018did} by incorporating tweets related to ten additional diseases, along with figurative mentions as a new label. They leveraged sentiment information and contextual embeddings to achieve improved performance compared to previous methods.

Khan et al. \cite{khan2020improving} enhanced health mention classification by employing an XLNet-based language model, while Khan et al. \cite{khan2022performance} conducted a comparative analysis of various transformer-based language models, highlighting the superior performance of the RoBERTa \cite{liu2019roberta} model on the health mention classification task.

Luo et al. \cite{luo2022covid} introduced a COVID-19 tweets dataset and proposed a dual-CNN classifier, comprising an auxiliary network (A-Net) to address class imbalance issues in the dataset.

Naseem et al. \cite{naseem2022robust} utilized domain-specific language models and linguistic features to enhance text representation, employing a Bi-LSTM with an attention module focused on disease or symptom terms. They also introduced PHS-BERT \cite{naseem2022benchmarking}, a domain-specific language model, and evaluated its performance on various social media datasets.

Naseem et al. \cite{naseem2022identification} released the Reddit Health Mention Dataset (RHMD), consisting of Reddit posts related to fifteen diseases. They developed HMCNET, combining emotional and domain-specific representations to classify given posts.

Rahman et al. \cite{rahman2024depressionemo} introduced DepressionEmo, a multi-label classification dataset derived from Reddit posts focusing on eight emotions related to depression. They experimented with machine learning and deep learning models, achieving the highest performance with a BART \cite{lewis2019bart} classifier.

Haque et al. \cite{haque2021deep} released a dataset for distinguishing posts related to depression versus suicidal ideation. They proposed a deep learning model coupled with an unsupervised label correction method, demonstrating the effectiveness of their approach.

Garg et al. \cite{garg2023annotated} constructed an annotated dataset for identifying risk factors of mental disturbance using Reddit posts. They established baseline models based on transformer architecture as well as LSTM and GRU for future research.

Poswiata et al. \cite{poswiata2022opi} categorized social media posts into various depression severity levels using BERT \cite{devlin2018bert}, RoBERTa \cite{liu2019roberta}, and XLNet \cite{yang2019xlnet} models.

Ghosal et al. \cite{ghosal2023depression} utilized fastText embeddings for contextual analysis, followed by TF-IDF for term relevance, and applied an XGBoost classifier to detect mental illness from social media posts.

\subsection{Disease detection using Contrastive Adversarial Training}
Khan et al. \cite{khan2022improving} introduced the Contrastive Adversarial Training (CAT) method, utilizing Fast Gradient Sign Method (FGSM) \cite{goodfellow2014explaining} to generate adversarial examples. Their approach involved two forward passes during LM training: one with clean training examples and another with adversarial examples. Additionally, they employed a contrastive loss that incorporated sentence-level representations of both clean and adversarial examples to enhance word representations. The overall training objective aimed to minimize a weighted loss comprising two classification losses for clean and adversarial examples, respectively, along with a contrastive loss.

In another work, Khan et al. \cite{khan2022novel} proposed a CAT-based method by augmenting Gaussian noise with a mean of $0$ and a standard deviation of $1$ in the hidden state representations of LMs. They found that adding noise to earlier layers of models yielded better results than adding noise in the middle or later layers. Additionally, Khan et al. \cite{khan2023unique} introduced the Random Weighted Perturbation (RWP) method, which perturbed the parameters of LMs based on their parameter distribution. They jointly trained models with clean and perturbed parameters, incorporating a contrastive loss similar to that in Khan et al. \cite{khan2022improving}. Furthermore, they proposed a meta-predictor leveraging the predictive performance of five language models.

Naseem et al. \cite{naseem2024linguistic} proposed a self-training method to generate new training examples to address label scarcity. Additionally, they employed a supervised contrastive loss to learn distinguishable representations between literal and figurative usage of disease or symptom terms.

This paper introduces a novel method termed self-augmentation for generating adversarial examples. Drawing inspiration from prior research \cite{khan2022improving,khan2022novel,khan2023unique}, our approach involves concurrently training LMs with both clean and adversarial examples. Through extensive experimentation, we demonstrate that this approach consistently outperforms various baseline and SotA methods on benchmark datasets.
\section{Method}
Our proposed method is illustrated in Fig. \ref{fig:methodology1} and comprises three main components.

The first component utilizes a transformer-based model fine-tuned for text classification. This model converts input text into feature embeddings and categorizes them into predefined categories. The second component also employs a transformer-based model, utilizing the same input text as the first component. However, during its fine-tuning process, the intermediate layer representations are explicitly augmented with representations from the first component, resulting in the generation of an adversarial version. The third and final component involves a shared projection network and contrastive loss. This component aims to bring the input sample and its perturbed version closer together while simultaneously pushing away other samples within the batch of inputs.

Now, we provide a detailed description of each of these components:
\begin{figure*}
\includegraphics[width=\textwidth]{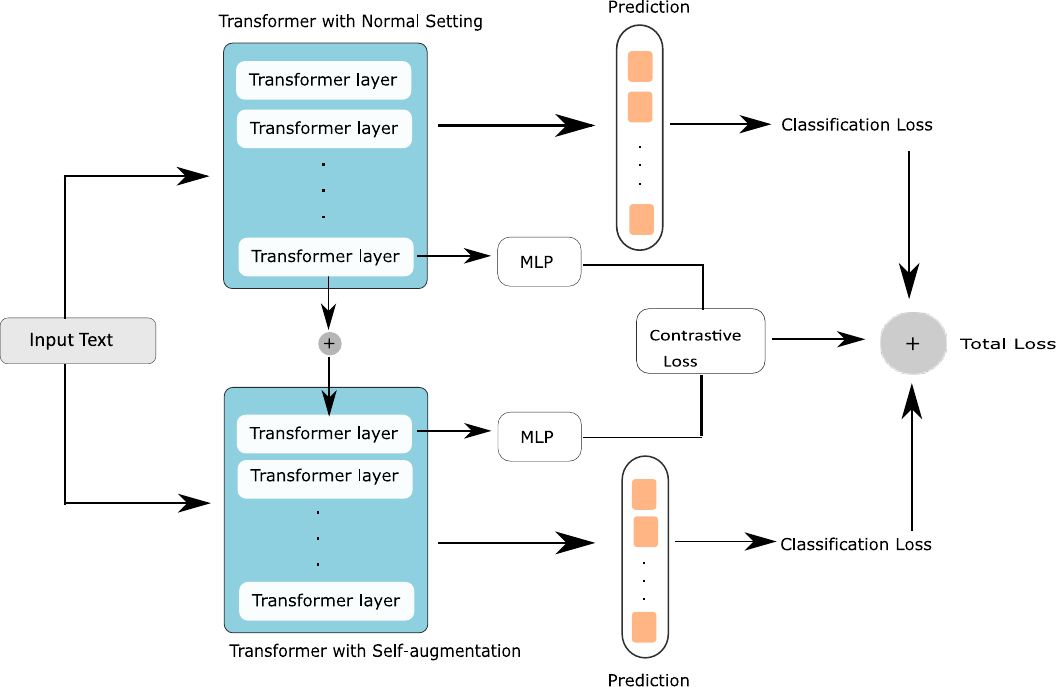}
\caption{The pipeline of our proposed method, consisting primarily of three components: a transformer model with standard settings, a transformer model with self-augmentation, and contrastive loss.} \label{fig:methodology1}
\end{figure*}
\subsection{Transformer Model}
Let $\digamma$ be a transformer model that serves as the backbone of our approach, responsible for mapping input tokens to embedding representations. We leverage transformer-based encoders such as BERT \cite{devlin2018bert} and RoBERTa \cite{liu2019roberta}. These models are fine-tuned on our datasets to learn representations for the input and minimize classification loss. The classification loss quantifies the disparity between the model's predictions, and the actual label. Cross-entropy is employed as the classification loss in our experiments.
\subsection{Self Augmentation}
The transformer model $\digamma$ comprises a series of L encoder layers, with each layer denoted as $L_i$ and its corresponding hidden representation as $H_i$ during the fine-tuning process. Additionally, let $C$ denote a copy of the model $\digamma$, with its layers represented as $L_j$  and hidden representations as $H_j$.

During the fine-tuning process of $\digamma$ and $C$, we implement self-augmentation by incorporating $H_i$ into the hidden representation $H_j$ of $C$. Mathematically, this augmentation is expressed as:

\begin{equation}
 H_j = H_j \oplus H_i
\end{equation}

Here, the symbol $\oplus$ represents element-wise summation. The term ``self-augmentation'' signifies that the augmentation originates from the model itself. Both models $\digamma$ and $C$ are trained concurrently with identical inputs, differing only in the augmentation of hidden representations in $C$. Subsequently, these models complete their forward passes and return two classification losses.
\subsection{Training Objective}
\subsubsection{Classification Loss}
We employ cross-entropy as the classification loss for both models, measuring the disparity between predictions and the actual label.
\subsubsection{Contrastive Loss}
Inspired by \cite{khan2022improving,khan2023unique}, our approach incorporates a contrastive learning method \cite{zbontar2021barlow} based on the redundancy injection principle. This method operates on two inputs: the clean input and its perturbed version, aiming to learn similar representations for corresponding pairs. Essentially, it learns a square matrix $M$ to depict the correlation between clean and perturbed inputs, where values range from -1 to +1, indicating anti-correlation and perfect correlation, respectively.

In our experiments, the inputs to the contrastive loss are denoted as $H_i$ and $H_j$. Following the methodology of prior studies \cite{khan2022improving,khan2023unique}, we pass $H_i$ and $H_j$ through a shared multi-layer perceptron (MLP) based projection network. This network projects them onto a similar representation space, subsequently passing them to the contrastive loss function. We adopt configurations for the projection network akin to those used in previous works \cite{khan2022improving,khan2023unique}.

The overall training loss minimized is a weighted sum of two classification losses and the contrastive loss, formulated as:

\begin{equation}
 L = \frac{(1 - \alpha)}{2}(L_{CE_1} + L_{CE_2}) + \alpha L_c
\end{equation}
Here, $L_{CE_1}$ and $L_{CE_2}$ denote two classification losses, and $L_c$ reprresents contrastive loss. The parameter $\alpha$ serves as a trade-off parameter, governing the weights assigned to the three losses. 
\section{Experiments}
In Table \ref{tbl:stats}, we present the statistics of the datasets used for training and evaluating our proposed method. Detail of each of the datasets is as follows:
\subsection{Benchmark Datasets}
\subsubsection{Dreaddit}
The dataset comprises posts sourced from the Reddit platform, covering a range of categories including anxiety, domestic violence, PTSD, relationships, stress, assistance for survivors of abuse, homelessness, food pantry, and almost homeless. Each post is labeled as either depression or non-depression. In total, the dataset contains 3,553 posts, with 2,838 posts allocated for training and 715 for testing. We divided the training set into 20\% validation set, and used remaining 80\% as training set.
\subsubsection{Reddit Health Mention Dataset (RHMD):}
The dataset comprises 10,015 Reddit posts covering 15 diseases, including asthma, OCD, diabetes, depression, cough, heart attack, PTSD, headache, cancer, allergy, Alzheimer's, stroke, addiction, migraine, and fever. Each post is labeled into one of four classes: personal health mention (PHM), non-personal health mention (NPHM), figurative mention (FM), and hyperbolic mention (HM). It's worth noting that we utilized a publicly available version of the dataset where HM and FM are combined. We divided dataset into  70\%  training set and 15\% each of the validation and test sets.
\subsubsection{DepressionEmo:}
The dataset is a multi-label dataset comprising 6,037 Reddit posts, each associated with one or more of 8 depression-related emotions: anger, cognitive dysfunction, emptiness, hopelessness, loneliness, sadness, suicide intent, and worthlessness. The distribution of posts across these emotions is as follows: 21.2\% for sadness, 19.1\% for hopelessness, 13.6\% for worthlessness, 12.7\% for loneliness, 11.1\% for anger, 10.3\% for emptiness, 6.8\% for suicide intent, and 5.2\% for cognitive dysfunction. For model training and evaluation, the dataset is divided into 4,225 training examples, 906 validation examples, and 906 test examples.
\begingroup
\setlength{\tabcolsep}{8pt} 
\renewcommand{\arraystretch}{1.3}
\begin{table*}[!htbp]\centering
\caption{Statistics of Benchmark Datasets. BC, MC, and ML denote binary classification, multi-class classification, and multi-label classification tasks, respectively.}
\label{tbl:stats}
\begin{tabular}{llll}
\toprule
\hline
\multicolumn{1}{c}{Dataset}      & \multicolumn{1}{c}{No. of Samples} & \multicolumn{1}{c}{Type} & {No. of Classes/Labels} \\
\hline

\multicolumn{1}{c}{Dreaddit \cite{turcan2019dreaddit}}        & \multicolumn{1}{c}{3,553}    & \multicolumn{1}{c}{BC}                    & \multicolumn{1}{c}{2}              \\

\hline

\multicolumn{1}{c}{RHMD \cite{naseem2022identification}}        & \multicolumn{1}{c}{10,015}      & \multicolumn{1}{c}{MC}                & \multicolumn{1}{c}{3}              \\

\hline
\multicolumn{1}{c}{DepressionEmo \cite{rahman2024depressionemo}} & \multicolumn{1}{c}{6,037}     & \multicolumn{1}{c}{ML}     &  \multicolumn{1}{c}{8}         
\\
\hline
\bottomrule
\end{tabular}
\end{table*}
\endgroup
\subsection{Training details}
To determine the optimal hyperparameters for our proposed method, we conducted a grid search. Specifically, we experimented with varying batch size $b \in \{16,32\}$ and $\alpha \in \{0.1,0.2,0.3,0.4,0.5\}$. Additionally, we explored the selection of $L_i$ and $L_j$ from the range of $\{0,3,6,9,12,15,18,21\}$. For optimization, we employed the Adam optimizer \cite{kingma2014adam}. The model was trained for 20 epochs with a fixed learning rate of $1e^{-5}$, and training was halted if no improvement in F1-score on the validation set was observed for 5 consecutive epochs. We also extracted the $H_i$ and $H_j$ hidden states and projected them to lower dimensions, using a 3-layered shared projection network. This network consisted of the first two hidden layers with a size of 1024 each, followed by a final layer with a size of 300. ReLU was employed as the non-linear activation function, and 1-D batch normalization was applied after each layer. Both projected representations were passed to the contrastive loss function. Following prior works \cite{khan2022improving,khan2022novel,khan2023unique}, default parameters were utilized for the contrastive loss function.
\subsection{Results}
\begin{figure}
\begin{subfigure}{0.5\textwidth}
  \centering
  \includegraphics[width=1.0\linewidth]{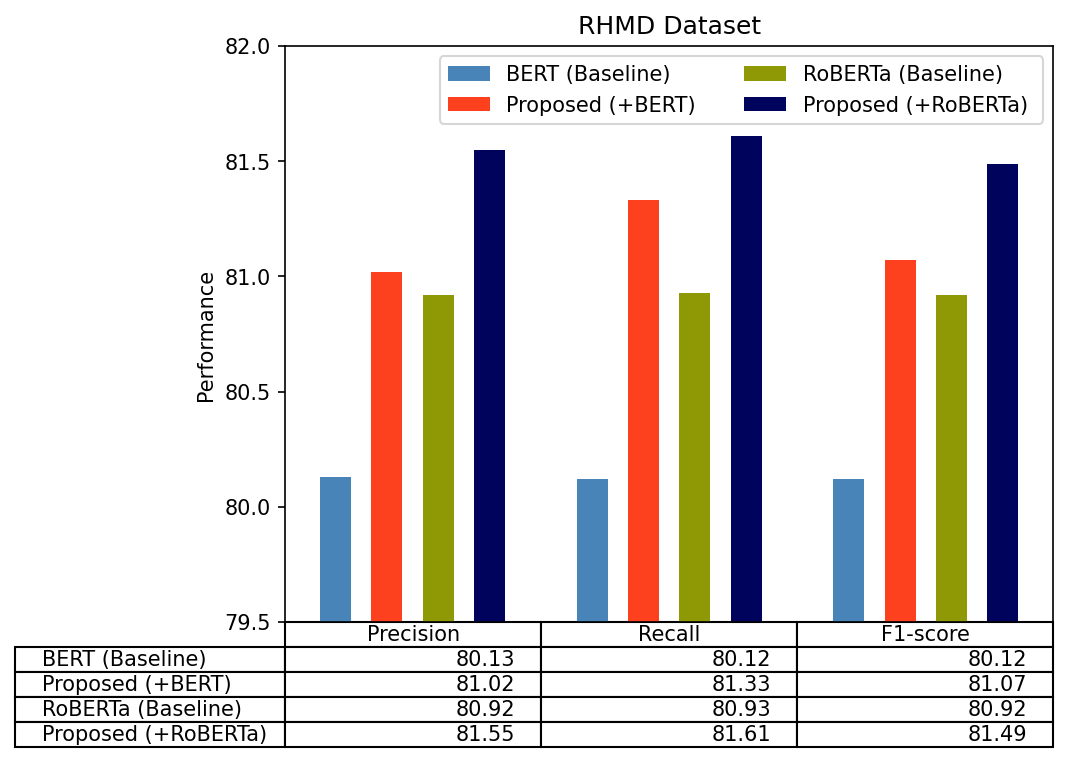}  
  \caption{Baseline vs. Proposed (RHMD).}
  \label{fig:rhmd1}
\end{subfigure}
\begin{subfigure}{.5\textwidth}
  \centering
  \includegraphics[width=1.0\linewidth]{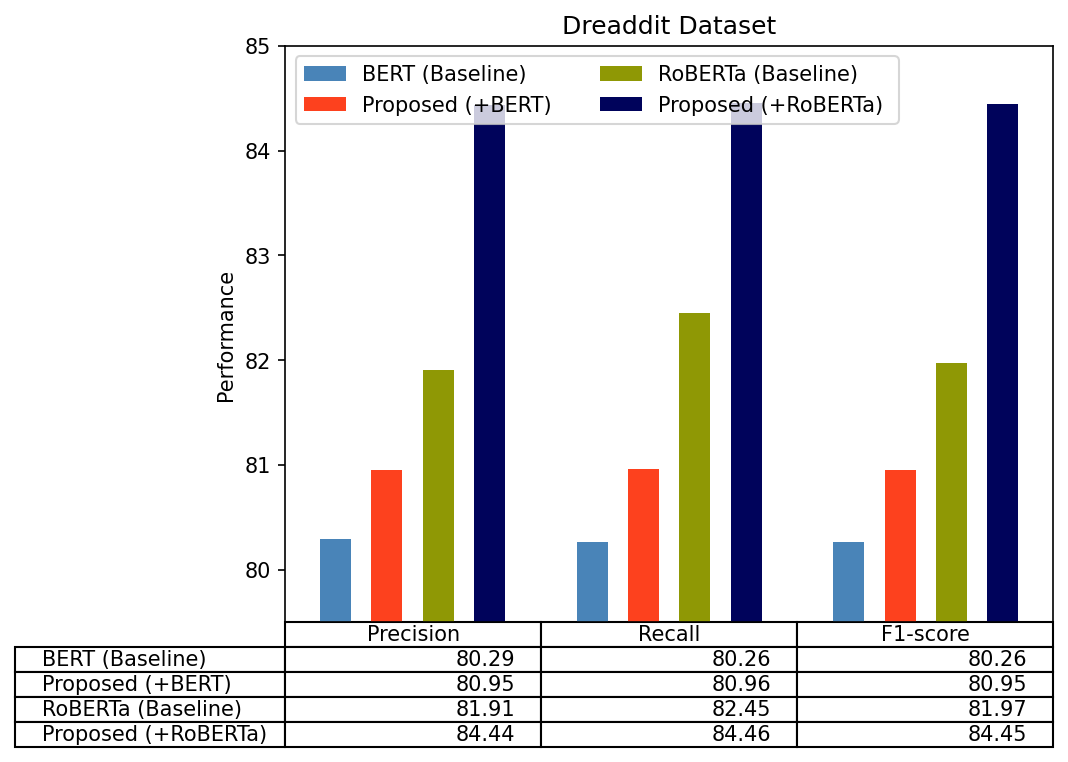}  
  \caption{Baseline vs. Proposed (Dreaddit).}
  \label{fig:dreaddit1}
\end{subfigure}

\begin{subfigure}{.5\textwidth}
  \centering
  \includegraphics[width=0.95\linewidth]{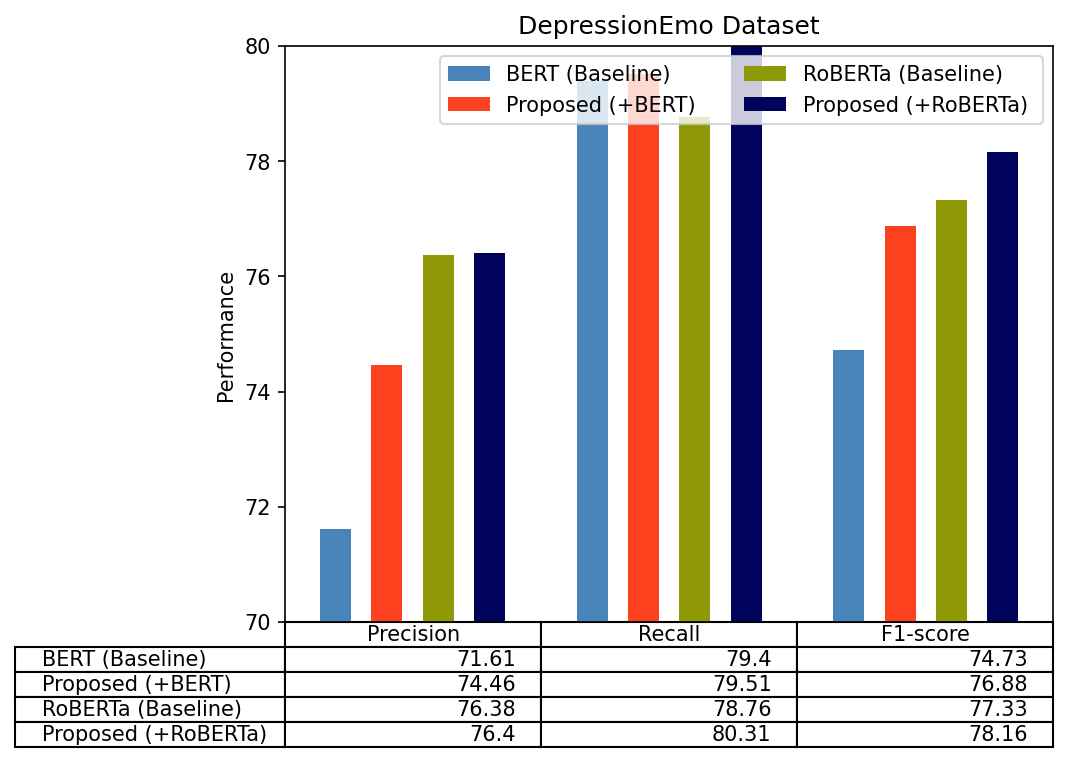}  
  \caption{Baseline vs. Proposed (DepressionEmo).}
  \label{fig:depemo1}
\end{subfigure}

\caption{Performance Comparison of the Proposed Approach with Baseline Methods across 3 Datasets.}
\label{fig:results-baseline}
\end{figure}
\begingroup
\setlength{\tabcolsep}{8pt} 
\renewcommand{\arraystretch}{1.3}

\begin{table*}[!htbp]\centering
\caption{Comparison of Proposed Method with SotA Results in terms of Precision, Recall, and F1-Scores, Reported on Test Sets.} 
\label{tbl:other-sota}
\begin{tabular}{|l|l|l|l|l|l} 
\cline{1-5}
Dataset                        & Method              & Precision      & Recall         & F1-score       &  \\ \cline{1-5}
\multirow{5}{*}{Dreaddit \cite{turcan2019dreaddit}}      & MentalBERT \cite{ji2021mentalbert}         & -              & 81.82          & 81.82          &  \\ \cline{2-5}

                               & Alghamdi et al. \cite{alghamdi2023studying}    & -              & -              & 82.88          &  \\ 
                               \cline{2-5}
                               & PHS-BERT \cite{naseem2022benchmarking}           & -              & -              & 82.89          &  \\ 
                               \cline{2-5}
                               & KC-Net \cite{yang2022mental}                & 84.1           & 83.3           & 83.5           &  \\ 
                               \cline{2-5}
                               & Proposed (+RoBERTa) & \textbf{84.44} & \textbf{84.46} & \textbf{84.45} &  \\ \cline{1-5}
\multirow{7}{*}{DepressionEmo \cite{rahman2024depressionemo}} & SVM \cite{rahman2024depressionemo}                 & 72.0           & 41.0           & 47.0           &  \\ \cline{2-5}
                               & Light GBM \cite{rahman2024depressionemo}           & 48.0           & 80.0           & 58.0           &  \\ \cline{2-5}
                               & XGBoost \cite{rahman2024depressionemo}             & 63.0           & 56.0           & 59.0           &  \\ \cline{2-5}
                               & GAN-BERT \cite{rahman2024depressionemo}            & 69.0           & 72.0           & 70.0           &  \\ \cline{2-5}
                               & BERT \cite{rahman2024depressionemo}           & 72.0           & 77.0           & 74.0           &  \\ \cline{2-5}
                               & BART \cite{rahman2024depressionemo}           & 70.0      &      \textbf{81.0}           & 76.0           &  \\ \cline{2-5}
                               & Proposed (+RoBERTa) & \textbf{76.4}           & 80.31          & \textbf{78.16}          &  \\ \cline{1-5}
\end{tabular}
\end{table*}
\endgroup
In Fig.\ref{fig:results-baseline}, we present a comparative analysis of the performance of our proposed method against baseline models across three datasets. As illustrated in Fig.\ref{fig:rhmd1}, our proposed method demonstrates a noteworthy improvement over the BERT baseline, achieving a 0.95\% increase in F1-score on the RHMD dataset. Additionally, when coupled with RoBERTa, our proposed method achieves an F1-score of 81.49\%, surpassing its baseline by 0.57\%. Fig.\ref{fig:dreaddit1} depicts the performance on the Dreaddit dataset, where the BERT baseline model achieves an F1-score of 80.26\%. In contrast, our proposed method outperforms the baseline by 0.69\%, while the performance is further elevated to 3.48\% with RoBERTa. In Fig.\ref{fig:depemo1}, the visualization showcases the performance of our proposed method on the DepressionEmo dataset, exhibiting improvements of 2.15\% and 0.83\% in F1-score over the respective baselines. These results strongly validate the superiority of our proposed method.

Table \ref{tbl:other-sota} provides a detailed performance comparison of our method with the SotA on the Dreaddit and DepressionEmo datasets. On the Dreaddit dataset, our proposed method with RoBERTa achieves a notable F1-score improvement of 0.95\% over the SotA, primarily enhancing recall by 1.16\%. Similarly, on the DepressionEmo dataset, our method achieves an F1-score of 78.16\%, surpassing the existing SotA by 2.16\%. While our method exhibits higher recall than precision, akin to existing deep learning based methods, it significantly enhances precision by 6.4\%.

Furthermore, to provide a comprehensive comparison with existing literature, we report average results over 10-fold cross-validations on the RHMD dataset in Table \ref{tbl:rhmd-sota}. Our proposed method with RoBERTa achieves the second-best results compared to the SotA methods. Although the highest-performing method \cite{khan2023unique} involves an ensemble of 5 trained models, resulting in increased inference time, our method slightly outperforms the model by Khan et al. \cite{khan2022improving}. However, its margin of improvement is significantly higher than other methods in the literature.

\begingroup
\setlength{\tabcolsep}{8pt} 
\renewcommand{\arraystretch}{1.3}
\begin{table*}[!htbp]\centering
\caption{Proposed vs. SotA results in terms of Precision, Recall, and  F1-scores for RHMD dataset. Results are averaged across 10-folds.} 
\label{tbl:rhmd-sota}
\begin{tabular}{llll}
\toprule
\hline
 \multicolumn{1}{c}{Model}   & \multicolumn{1}{c}{Precision} & \multicolumn{1}{c}{Recall}   & \multicolumn{1}{c}{F1-score} \\
\toprule
\hline
\multicolumn{1}{c}{BERT \cite{devlin2018bert}}   & \multicolumn{1}{c}{63.0}    & \multicolumn{1}{c}{68.0} & \multicolumn{1}{c}{65.0}    \\
\hline
 \multicolumn{1}{c}{BioBERT \cite{lee2020biobert}} & \multicolumn{1}{c}{65.0}    & \multicolumn{1}{c}{62.0}   & \multicolumn{1}{c}{63.0}    \\
\hline
 \multicolumn{1}{c}{CT-BERT \cite{muller2023covid}}   & \multicolumn{1}{c}{65.0}    & \multicolumn{1}{c}{68.0} & \multicolumn{1}{c}{67.0}   \\
 \hline
 \multicolumn{1}{c}{PHS-BERT \cite{naseem2022benchmarking}}   & \multicolumn{1}{c}{67.0}    & \multicolumn{1}{c}{69.0} & \multicolumn{1}{c}{68.0}   \\
 \hline

 \multicolumn{1}{c}{WESPAD \cite{karisani2018did}}   & \multicolumn{1}{c}{60.0}    & \multicolumn{1}{c}{60.0} & \multicolumn{1}{c}{59.0}   \\
 \hline
\multicolumn{1}{c}{FeatAug+ \cite{iyer2019figurative}}   & \multicolumn{1}{c}{51.0}    & \multicolumn{1}{c}{51.0} & \multicolumn{1}{c}{51.0}   \\
\hline
\multicolumn{1}{c}{JiangLSTM \cite{jiang2018identifying}}   & \multicolumn{1}{c}{63.0}    & \multicolumn{1}{c}{63.0} & \multicolumn{1}{c}{63.0}   \\
\hline
\multicolumn{1}{c}{BERT-MTL \cite{aduragba2023improving}}   & \multicolumn{1}{c}{69.0}    & \multicolumn{1}{c}{65.0} & \multicolumn{1}{c}{67.0}   \\
\hline
\multicolumn{1}{c}{BiLSTM-Senti \cite{biddle2020leveraging}}   & \multicolumn{1}{c}{67.0}    & \multicolumn{1}{c}{68.0} & \multicolumn{1}{c}{68.0}   \\
\hline
\multicolumn{1}{c}{BiLSTM-Attn+Senti \cite{naseem2022robust} }   & \multicolumn{1}{c}{70.0}    & \multicolumn{1}{c}{71.0} & \multicolumn{1}{c}{71.0}   \\
\hline
\multicolumn{1}{c}{HMCNET \cite{naseem2022identification}}   & \multicolumn{1}{c}{75.0}    & \multicolumn{1}{c}{75.0} & \multicolumn{1}{c}{75.0}   \\
\hline
\multicolumn{1}{c}{Naseem et al. (+BERT) \cite{naseem2024linguistic}}   & \multicolumn{1}{c}{78.0}    & \multicolumn{1}{c}{79.0} & \multicolumn{1}{c}{79.0}   \\
\hline
\multicolumn{1}{c}{Naseem et al. (+PHS-BERT) \cite{naseem2024linguistic}}   & \multicolumn{1}{c}{81.0}    & \multicolumn{1}{c}{81.0} & \multicolumn{1}{c}{81.0}   \\
\hline

\multicolumn{1}{c}{Khan et al. \cite{khan2022improving}}   & \multicolumn{1}{c}{83.43}    & \multicolumn{1}{c}{83.27} & \multicolumn{1}{c}{83.23}   \\
\hline

\multicolumn{1}{c}{Khan et al. \cite{khan2023unique}}   & \multicolumn{1}{c}{\textbf{83.82}}    & \multicolumn{1}{c}{\textbf{83.91}} & \multicolumn{1}{c}{\textbf{83.8}}   \\
\hline 

\multicolumn{1}{c}{Proposed (+RoBERTa)}   & \multicolumn{1}{c}{83.40}    & \multicolumn{1}{c}{83.29} & \multicolumn{1}{c}{83.3}   \\
\hline
\bottomrule
\end{tabular}
\end{table*}
\endgroup

\begingroup
\setlength{\tabcolsep}{8pt} 
\renewcommand{\arraystretch}{1.3}

\subsection{Analysis of the Proposed Method}

\begingroup
\setlength{\tabcolsep}{8pt} 
\renewcommand{\arraystretch}{1.3}
\begin{table*}[!htbp]\centering
\caption{Optimal Layer Combination and Weighting Parameter $\alpha$ on Validation Sets for 3 Datasets.} 
\label{tbl:hp}
\begin{tabular}{|l|l|l|l|l|l|l}
\cline{1-6}
Dataset                        & Model              & $\alpha$ & $L_i$ & $L_j$ & F1-score &  \\ \cline{1-6}
\multirow{2}{*}{Dreaddit \cite{turcan2019dreaddit}}      & Proposed(+BERT)    & 0.1   & 18 & 21 & 84.76    &  \\ \cline{2-6}
                               & Proposed(+RoBERTa) & 0.1   & 18 & 3  & 86.56    &  \\ \cline{1-6}
\multirow{2}{*}{DepressionEmo \cite{rahman2024depressionemo}} & Proposed(+BERT)    & 0.2   & 18 & 0  & 78.90    &  \\ \cline{2-6}
                               & Proposed(+RoBERTa) & 0.5   & 21 & 9  & 80.08    &  \\ \cline{1-6}
\multirow{2}{*}{RHMD \cite{naseem2022identification}}          & Proposed(+BERT)    & 0.4   & 9  & 21 & 82.52    &  \\ \cline{2-6}
                               & Proposed(+RoBERTa) & 0.4   & 0  & 15 & 84.44    &  \\ \cline{1-6}
\end{tabular}
\end{table*}
\endgroup
\subsubsection{Effect of Layer Combination}
In Table \ref{tbl:hp}, we present the layer combinations and the effect of $\alpha$ for our proposed method across three datasets. Notably, on the Dreaddit dataset, combining later layers from both models yields favorable results for the proposed method leveraging BERT. Conversely, the proposed method employing RoBERTa performs well when the later layer is combined with an earlier layer. The optimal value of $\alpha$ for both models is found to be 0.1. For the DepressionEmo dataset, combining later layers from the first model with later layers from the second model proves effective, although the optimal value of $\alpha$ varies for each model. On the other hand, for the RHMD dataset, combining earlier layers from the first model with later layers from the second model demonstrates superior performance, with the optimal value of $\alpha$ identified as 0.4 for both models. A notable trend observed in the Dreaddit and DepressionEmo datasets is that later layers from the first model enhance performance when combined with the second model. One possible explanation for this phenomenon is the limited number of training samples in these datasets, posing challenges to the learning capability of the models. Leveraging already well-learned semantic representations from later layers improves syntactic features for earlier layers.

Conversely, the RHMD dataset contains relatively larger training examples. Here, the combination of syntactic features learned in earlier layers with semantic features from later layers enhances overall performance.

\subsubsection{Ablation Study}
To analyze the effect of various components of the proposed
method, we conduct ablation studies. It is clear from Table \ref{tbl:ablation} that the F1-score for Dreaddit dataset is decreased by 1.9\% if CL component is dropped.
The F1-score is further decreased by 0.58\% if self-augmentation component is
dropped. Overall F1-score is reduced by 2.48\% if the proposed components are
dropped. We observe different pattern for DepressionEmo dataset. The addition of self-augmentation actually decreases the F1-score over RoBERTa baseline by 0.38\%. One possible explanation for this behaviour is the complexity
of dataset itself. This complexity comes from the smaller sample size in the
dataset. Multi-labeling makes it further complex as unique combination of labels is small. However, adding CL component improves the text representation,
and F1-score is increased by 0.83\% compared to RoBERTa baseline. For RHMD
dataset, incorporating self-augmentation causes a minor improvement of F1-score by 0.12\%.
Possible justification of this minute difference is the complexity of the dataset
due to various number of diseases, i.e., 15 in the dataset. Hence, per-disease samples are small as compared to overall dataset. However similar to DepressionEmo
dataset, adding CL causes increase in F1-score over the BERT baseline.
\subsubsection{Visualization of Text Representations}
In Fig. \ref{fig:embedd}, embedding are visualized for Dreaddit and DepressionEmo datasets. For dreaddit dataset, our proposed method with BERT learns different representations from its baseline by effectively separating depression and non-depression classes. On the other hand, proposed model with RoBERTa spreads the representations while maintaining the decision boundary between two classes. For the DepressionEmo dataset, our proposed method demonstrates embeddings with a smaller spread compared to the baseline methods. Furthermore, the representations generated by our proposed methods exhibit numerous vacant spaces in the representation space, indicating a stronger ability to learn decisive boundaries between classes.
Similarly, in Fig. \ref{fig:embedd-rhmd}, the representations learned by our proposed methods are notably more compact, with embeddings of various classes distinctly spaced apart from one another.

\begingroup
\setlength{\tabcolsep}{8pt} 
\renewcommand{\arraystretch}{1.3}

\begin{table*}[!htbp]\centering
\caption{Ablation Study: Analyzing the Impact of Adding Self-augmentation (SA) and Contrastive Learning (CL) on Test Sets.} 
\label{tbl:ablation}
\begin{tabular}{|l|l|l|l|l|l}
\cline{1-5}
Dataset                        & Method            & Precision & Recall & F1-score &  \\ \cline{1-5}
\multirow{3}{*}{Dreaddit \cite{turcan2019dreaddit}}      & RoBERTa (Baseline)          & 81.91     & 82.45  & 81.97    &  \\ \cline{2-5}
                               & RoBERTa (+SA)     & 82.49     & 82.94  & 82.55    &  \\ \cline{2-5}
                               & RoBERTa (+Proposed) & 84.44     & 84.46  & 84.45    &  \\ \cline{1-5}
\multirow{3}{*}{DepressionEmo \cite{rahman2024depressionemo}} & RoBERTa (Baseline)         & 76.38     & 78.76  & 77.33    &  \\ \cline{2-5}
                               & RoBERTa(+SA)      & 73.99     & 80.50  & 76.95    &  \\ \cline{2-5}
                               & RoBERTa (+Proposed) & 76.40     & 80.31  & 78.16    &  \\ \cline{1-5}
\multirow{3}{*}{RHMD \cite{naseem2022identification}}          & BERT (Baseline)              & 80.13     & 80.12  & 80.12    &  \\ \cline{2-5}
                               & BERT (+SA)          & 80.16     & 80.40  & 80.24    &  \\ \cline{2-5}
                               & BERT (+Proposed)    & 81.02     & 81.33  & 81.07    &  \\ \cline{1-5}
\end{tabular}
\end{table*}
\endgroup

\begin{figure}
\begin{subfigure}{0.5\textwidth}
  \centering
  \includegraphics[width=1.0\linewidth]{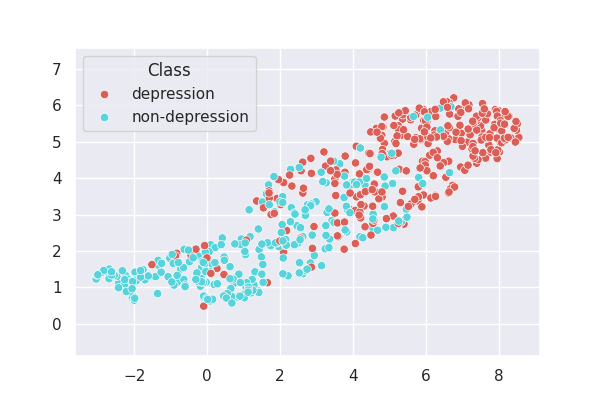}  
  \caption{BERT baseline (Dreaddit).}
  \label{fig:rhmd}
\end{subfigure}
\begin{subfigure}{.5\textwidth}
  \centering
  \includegraphics[width=1.0\linewidth]{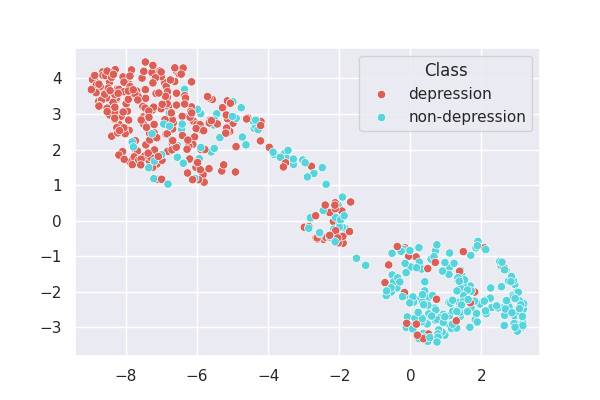}  
  \caption{BERT Proposed (Dreaddit).}
  \label{fig:dreaddit}
\end{subfigure}

\begin{subfigure}{.5\textwidth}
  \centering
  \includegraphics[width=1.0\linewidth]{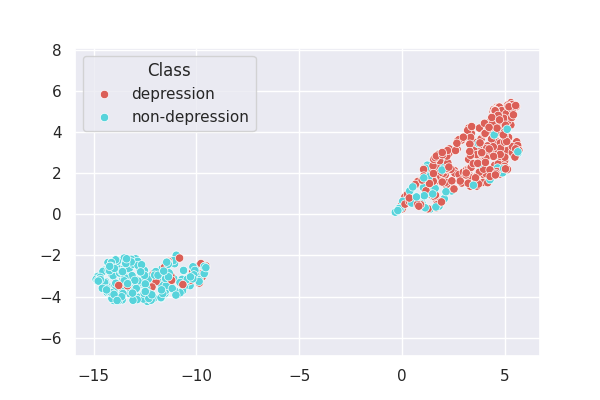}  
  \caption{RoBERTa baseline (Dreaddit).}
  \label{fig:depemo}
\end{subfigure}
\begin{subfigure}{.5\textwidth}
  \centering
  \includegraphics[width=1.0\linewidth]{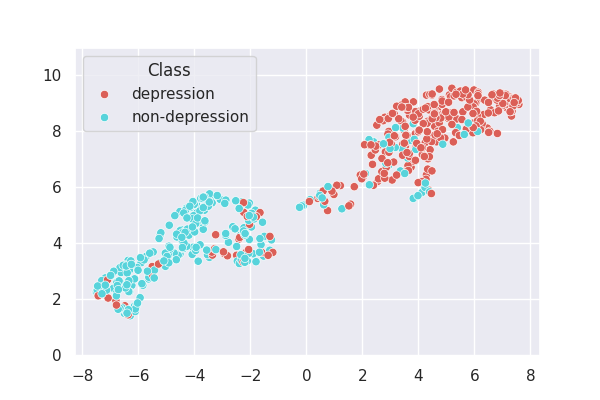}  
  \caption{RoBERTa Proposed (Dreaddit).}
  \label{fig:depemo}
\end{subfigure}

\begin{subfigure}{0.5\textwidth}
  \centering
  \includegraphics[width=1.0\linewidth]{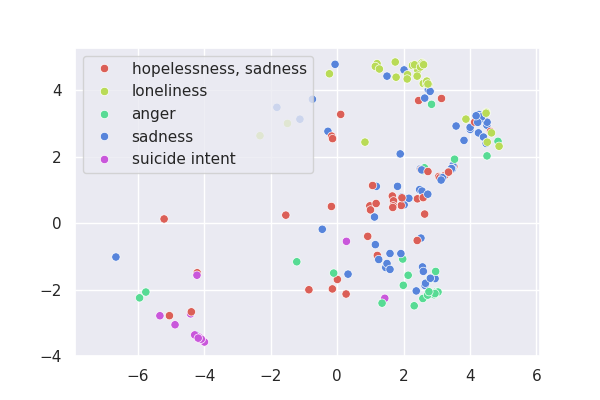}  
  \caption{BERT baseline (DepressionEmo).}
  \label{fig:rhmd}
\end{subfigure}
\begin{subfigure}{.5\textwidth}
  \centering
  \includegraphics[width=1.0\linewidth]{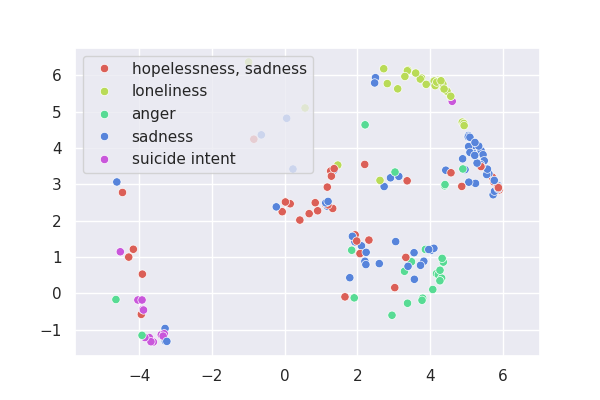}  
  \caption{BERT Proposed (DepressionEmo).}
  \label{fig:dreaddit}
\end{subfigure}

\begin{subfigure}{0.5\textwidth}
  \centering
  \includegraphics[width=1.0\linewidth]{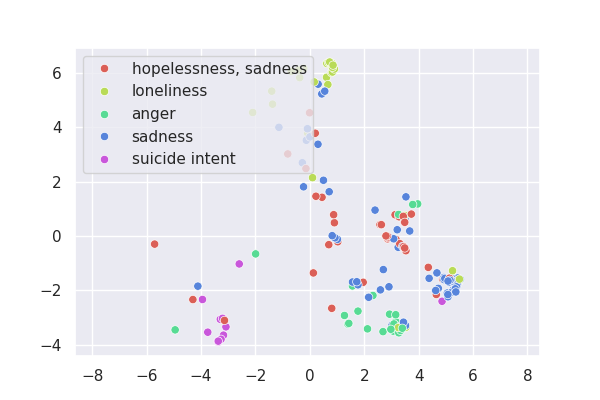}  
  \caption{RoBERTa baseline (DepressionEmo).}
  \label{fig:rhmd}
\end{subfigure}
\begin{subfigure}{.5\textwidth}
  \centering
  \includegraphics[width=1.0\linewidth]{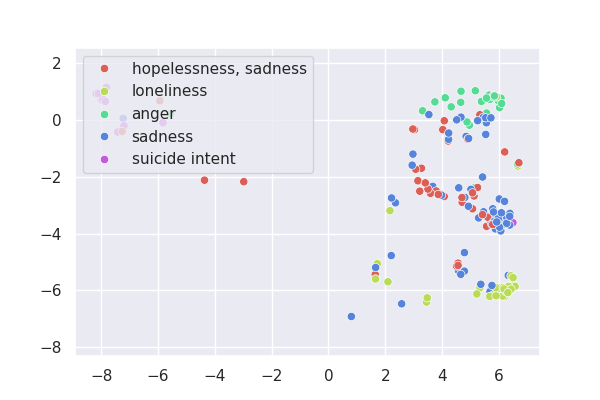}  
  \caption{RoBERTa Proposed (DepressionEmo).}
  \label{fig:dreaddit}
\end{subfigure}

\caption{Embedding Visualizations for Dreaddit and DepressionEmo Datasets. The embeddings are visualized for the entire validation set of the Dreaddit dataset, and a subset (for clearer visualization) of the DepressionEmo dataset.}
\label{fig:embedd}
\end{figure}

\begin{figure}
\begin{subfigure}{0.5\textwidth}
  \centering
  \includegraphics[width=1.0\linewidth]{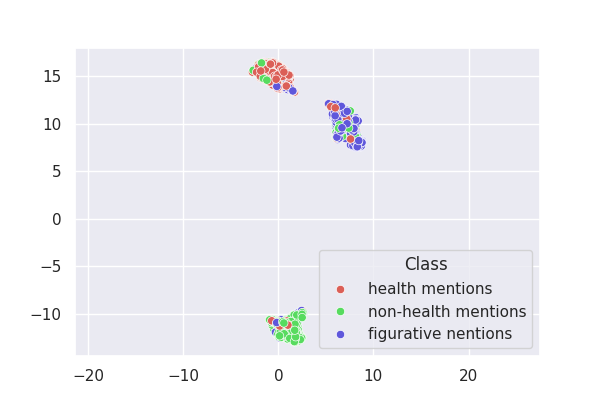}  
  \caption{BERT baseline.}
  \label{fig:rhmd}
\end{subfigure}
\begin{subfigure}{.5\textwidth}
  \centering
  \includegraphics[width=1.0\linewidth]{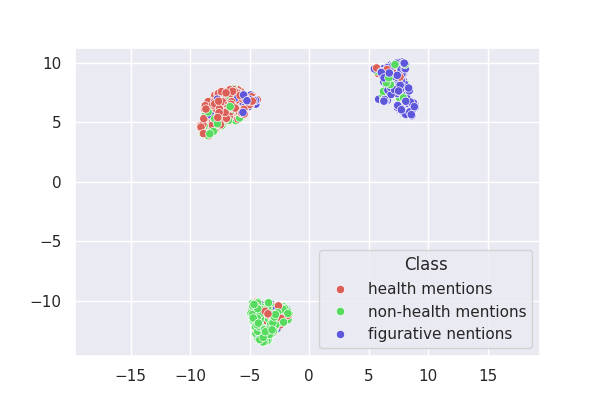}  
  \caption{BERT Proposed.}
  \label{fig:dreaddit}
\end{subfigure}

\begin{subfigure}{.5\textwidth}
  \centering
  \includegraphics[width=1.0\linewidth]{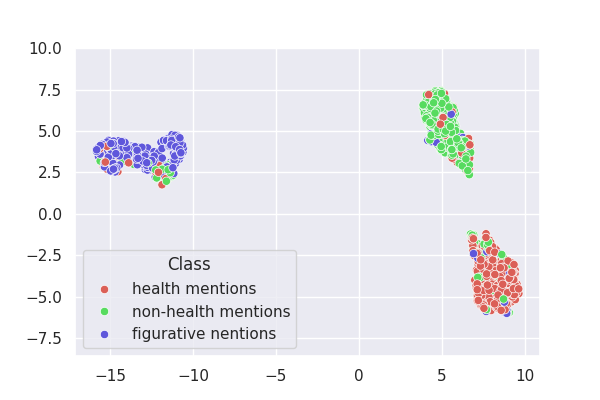}  
  \caption{RoBERTa baseline.}
  \label{fig:depemo}
\end{subfigure}
\begin{subfigure}{.5\textwidth}
  \centering
  \includegraphics[width=1.0\linewidth]{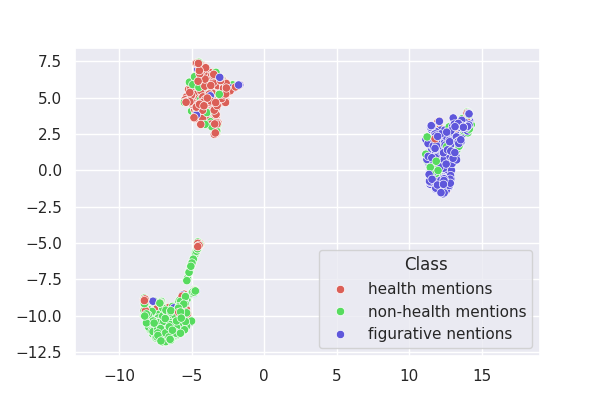}  
  \caption{RoBERTa Proposed.}
  \label{fig:depemo}
\end{subfigure}

\caption{Embedding Visualizations for RHMD Dataset.}
\label{fig:embedd-rhmd}
\end{figure}

\section{Conclusion}
In this paper, we have advanced the field of disease detection from social media text by improving the text representations capabilities of LMs by proposing a new method. We proposed self-augmentation technique for generating adversarial examples and then unified adversarial training with contrastive learning methods, where it demonstrated notable improvements in models performance. Through extensive experiments across 3 benchmark datasets comprising diverse types of classification tasks on social media text, we have demonstrated the efficacy of our proposed approach. Our method has achieved SotA performance on these datasets by enhancing discriminative features of LMs. Our future investigations will focus on evaluating the scalability and performance of this method on LMs with billions of parameters. 
%
%
%
%
\bibliographystyle{splncs04}
\bibliography{references}
\end{document}